\documentclass[twocolumn,dvipsnames,table]{article}

\usepackage{reportstyle}
\usepackage[utf8]{inputenc}
\usepackage[T1]{fontenc}

\usepackage[labelfont=bf]{caption}
\usepackage{sectsty}
\usepackage{fancyhdr}
\let\cite\citep
\renewcommand{\headrulewidth}{1pt}
\makeatletter
\def\headrule{{\if@fancyplain\let\headrulewidth\plainheadrulewidth\fi
  \hrule\@height\headrulewidth\@width\headwidth\vskip-\headrulewidth}}
\makeatother
\AtBeginDocument{\setlength{\headwidth}{\textwidth}}

\definecolor{HYDarkBlue}{HTML}{2155EA}
\definecolor{HYLightBlue}{HTML}{A8DFF6}
\usepackage[
  colorlinks,
  linkcolor=HYDarkBlue,
  anchorcolor=HYDarkBlue,
  citecolor=HYDarkBlue,
  urlcolor=HYDarkBlue
]{hyperref}
\makeatletter\g@addto@macro\UrlBreaks{\do\0\do\1\do\2\do\3\do\4\do\5\do\6\do\7\do\8\do\9\do\a\do\b\do\c\do\d\do\e\do\f}\makeatother
\sectionfont{\color{HYDarkBlue}\fontfamily{zi4}\selectfont}
\subsectionfont{\color{HYDarkBlue}\fontfamily{zi4}\selectfont}
\subsubsectionfont{\color{HYDarkBlue}\fontfamily{zi4}\selectfont}
\usepackage{url}
\usepackage{booktabs}
\usepackage{amsmath}
\usepackage{amssymb}
\usepackage{graphicx}
\usepackage{xcolor}
\usepackage{xspace}
\usepackage{enumitem}
\usepackage{pifont}
\newcommand{\cmark}{\ding{51}}
\newcommand{\xmark}{\ding{55}}

\newcommand{\mytitle}{HyScale: Scaling Long-Context Training for Learned Sparse Attention}
\newcommand{\sysname}{HyScale\xspace}
\newcommand{\flashtopk}{FlashTopK\xspace}
\newcommand{\hypreview}{Hy4 preview\xspace}
\newcommand{\waste}[1]{\textcolor[HTML]{C0392B}{\textcircled{\scriptsize\textbf{#1}}}}
\newcommand{\fix}[1]{\textcolor[HTML]{2C5D96}{\textcircled{\scriptsize\textbf{#1}}}}
\newcommand{\sparsereplay}{selection replay\xspace}
\newcommand{\overlap}{decoupled CP overlap\xspace}
\newcommand{\Sparsereplay}{Selection replay\xspace}
\newcommand{\Overlap}{Decoupled CP overlap\xspace}
\newcommand{\indexreplay}{index replay\xspace}
\newcommand{\scorereplay}{score replay\xspace}
\newcommand{\Indexreplay}{Index replay\xspace}
\newcommand{\Scorereplay}{Score replay\xspace}

\title{\mytitle}

\author{%
  \textbf{\small Yangshen Deng$^{1}$, \quad Xinjing Huang$^{2}$, \quad Yulong Wang$^{2}$, \quad Yunzhuo Liu$^{2}$, \quad Jun Hu$^{2}$,}\\[0.2em]
  \textbf{\small Shuyong Tan$^{2}$, \quad Chun-Chih Tseng$^{2}$, \quad Mingji Han$^{2}$, \quad Ray Ying$^{2}$, \quad Key Zhang$^{2}$, \quad Luo Mai$^{1}$}\\[0.45em]
  {\normalfont\small $^{1}$University of Edinburgh \quad $^{2}$Tencent Hunyuan}
  \vspace{0em}
}

\hypersetup{
  pdftitle={HyScale: Scaling Long-Context Training for Learned Sparse Attention},
  pdfauthor={Yangshen Deng, Xinjing Huang, Yulong Wang, Yunzhuo Liu, Jun Hu, Shuyong Tan, Chun-Chih Tseng, Mingji Han,
    Ray Ying, Key Zhang, Luo Mai},
  pdfdisplaydoctitle=true
}

\newcommand{\emailicon}{%
  \raisebox{-1.5pt}{\includegraphics[height=1.05em]{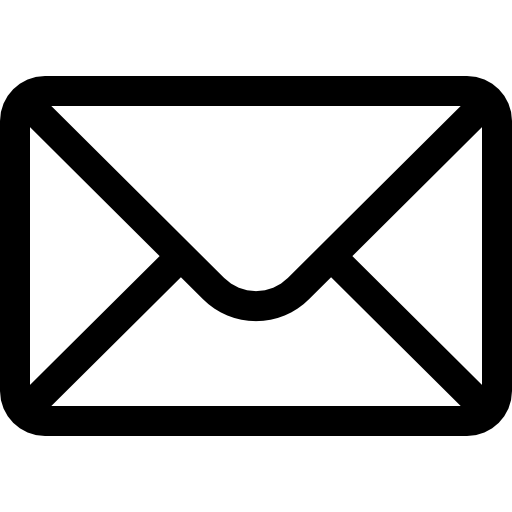}}}
\newcommand{\artifactlinks}{%
  \begin{tabular}{rl}
    \emailicon & \href{mailto:yangshen.deng@ed.ac.uk}{\nolinkurl{yangshen.deng@ed.ac.uk}}
  \end{tabular}%
}

\begin{document}

\maketitle
\thispagestyle{fancy}

\begin{abstract}
We present \sysname, a long-context training system for learned sparse attention. Learned sparse attention is a promising architecture for million-token contexts, widely adopted in production since DeepSeek-V3.2. Selection, however, grows into the training bottleneck. For each query, it scores every previous token whereas sparse attention reads only $k$; worse, current systems materialize a full score table and repeat selection during recomputation. Our key insight is that the output of selection is not an activation, but rather a set of indices. Building on it, \sysname introduces \textit{\flashtopk} and \emph{\sparsereplay}. \textit{\flashtopk} finds each query's exact top-$k$ on the fly without a score table, while \emph{\sparsereplay} reuses forward-pass selection in recomputation. Together with holistic system optimizations tackling end-to-end challenges, \sysname speeds up GLM-5.1 training on 1M-token sequences by 2.07$\times$ over Megatron-LM on 512 H20 GPUs. It also speeds up the CSA layers of DeepSeek-V4-Flash and the QSA layers of Qwen3.8-Flash-Next by 2.01$\times$ and 1.57$\times$ at 1M tokens on 32 H800 GPUs.
\end{abstract}

\section{Introduction}
\label{sec:intro}
Long context has become a core capability of large language models, driving training sequence lengths from 128K to 1M tokens across long documents, code repositories, and agent trajectories. At these lengths, full attention dominates training cost: its computation grows quadratically with context length because each query attends to every previous token.
Learned sparse attention~\cite{nsa,dsv32} reduces this cost by splitting attention into two steps. In \emph{selection}, a lightweight indexer scores every previous token for each query and keeps the top-$k$. In \emph{sparse attention}, each query attends only to the selected tokens. Token-level designs, pioneered by DeepSeek Sparse Attention (DSA)~\cite{dsv32}, are widely adopted in production models such as GLM-5.1~\cite{glm5} and \hypreview~\cite{hy4preview}. Block-level variants select contiguous token blocks instead, including CSA in DeepSeek-V4~\cite{dsv4}, QSA in Qwen3.8-Flash-Next~\cite{qwen38next}, and MSA in MiniMax-M3~\cite{msa}.

Selection, however, grows into the new training bottleneck. While sparse attention reads only $k$ tokens per query, selection still scores every previous token. Worse, current systems amplify this cost in two ways. First, they \emph{materialize} a full score table across all previous tokens, even though top-$k$ keeps only $k$ entries per query. Second, they \emph{repeat} selection during activation recomputation~\cite{chen2016training,korthikanti2023reducing} by rerunning the entire layer from scratch. In Megatron-LM~\cite{shoeybi2019megatron}, selection consequently consumes 47\% of a GLM-5.1 training step at 1M tokens, excluding the pipeline wait—6.6$\times$ the time of sparse attention (\S\ref{sec:inefficiencies}).

Both inefficiencies stem from treating the output of selection as a standard activation. Our key insight is that \textbf{the output of selection is not a continuous activation, but rather a discrete set of indices}. Unlike standard activations, top-$k$ selection is non-differentiable: the backward pass requires neither the intermediate score table nor a rerun of top-$k$ search. Specifically, sparse attention uses only the selected indices, and the indexer loss differentiates through only the $k$ selected scores. Consequently, training can find these indices on the fly without materializing the full score table, and directly reuse them during recomputation instead of repeating selection.

Realizing this insight in end-to-end training raises three challenges. First, on-the-fly top-$k$ selection must decide whether to retain or discard each score immediately upon computation. Existing sampling-based top-$k$ operators~\cite{litetopk,dosa} estimate a score threshold for inference, but break down in training: their neighbor sampling is incompatible with context parallelism (CP), their sampling and key pruning both fail in early training, and selection becomes non-deterministic. Second, existing selective recomputation policies~\cite{korthikanti2023reducing,pytorchsac} operate at the granularity of whole operators, whereas the indexer loss requires only the $k$ selected scores. Consequently, recomputation must either retain the full score table or rescore all previous tokens merely to recover $k$ values per query.
Third, once selection is accelerated, CP communication and inefficient sparse kernels become exposed. Existing systems gather indexer keys and KV states synchronously before selection—stalling selection on KV transfers it never reads—while their sparse kernels assign each thread block a single query, repeating the loads and updates of entries adjacent queries share.

To address these challenges, we present \sysname{} (Figure~\ref{fig:arch}), a long-context training system for learned sparse attention. We make three contributions:

\begin{figure}[t]
\centering
\includegraphics[width=\columnwidth]{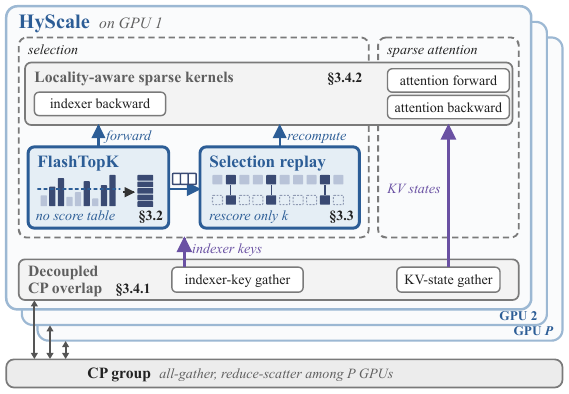}
\caption{\textbf{\sysname architecture.}}
\label{fig:arch}
\end{figure}

\paragraph{(1) \flashtopk, an on-the-fly top-$k$ selection operator via optimistic thresholding.} \flashtopk selects top-$k$ entries while scoring, eliminating the score table entirely. The key insight is that threshold-based filtering can tolerate optimistic thresholds as long as failures are rare and cheap to recover from. Built with optimistic strided sampling and adaptive key pruning, \flashtopk delivers fast, exact, and deterministic \mbox{top-$k$} selection with full CP compatibility.

\paragraph{(2) \Sparsereplay, a fine-grained recomputation policy for selection.} Rather than operating at the operator level, \sparsereplay records forward-pass selection results and replays only the $k$ selected entries during recomputation. This fine-grained replay eliminates repeated selection with negligible memory overhead, reducing recomputation cost by orders of magnitude over existing selective policies.

\paragraph{(3) \sysname, a holistic training system for learned sparse attention.}
Built on Megatron-LM, \sysname integrates \flashtopk and \sparsereplay with end-to-end system optimizations: \overlap hides the CP collectives of sparse attention behind selection, while locality-aware sparse kernels process two adjacent queries together so that they load and update their shared entries once. \sysname supports diverse learned sparse attention architectures, including DSA, CSA, and QSA.

Across 512 H20 GPUs, \sysname accelerates end-to-end GLM-5.1 training on 1M-token sequences by 2.07$\times$ over Megatron-LM, with speedups scaling further as context length grows. Beyond token-level DSA, \sysname generalizes seamlessly to block-level architectures: at 1M tokens on 32 H800 GPUs, it accelerates the CSA layers of DeepSeek-V4-Flash by 2.01$\times$ and the QSA layers of Qwen3.8-Flash-Next by 1.57$\times$. \sysname is deployed in production, powering large-scale LLM training at million-token contexts.

\section{Background and Motivation}
\label{sec:background}

\subsection{Learned Sparse Attention}
\label{sec:bg-lsa}

Learned sparse attention lets each query attend to only a few entries. An entry is an earlier token or a block of earlier tokens. Each layer runs in two steps. \emph{Selection} uses a learned indexer to score every earlier entry of a query and keeps the $k$ highest. \emph{Sparse attention} then reads only the KV states of the selected entries (Appendix~\ref{app:formulation}). Selection thus costs up to $O(L)$ in computation per query for $L$ tokens, while sparse attention costs only $O(k)$.

In training, gradients likewise reach only the selected entries. Top-$k$ is not differentiable. The indexer therefore learns from a loss of its own, defined on its $k$ selected scores. This loss pulls these scores toward the attention distribution over the selected entries. Sparse attention also back-propagates only to the selected entries (Appendix~\ref{app:backward}).

Designs differ mainly in what an entry is and how it is scored. DSA~\cite{dsv32} scores single tokens with a multi-head lightning indexer. All indexer heads share one indexer key per token. Its sparse attention uses multi-head latent attention (MLA)~\cite{deepseekv2}. All attention heads likewise share one compressed KV latent per token. Block-level designs score blocks of tokens instead. CSA in DeepSeek-V4~\cite{dsv4} and QSA in Qwen3.8-Flash-Next~\cite{qwen38next} compress or mean-pool every four tokens into one entry. MSA in MiniMax-M3~\cite{msa} takes the maximum token score within each block of 128 tokens.

\subsection{Long-Context Training}
\label{sec:bg-training}

\paragraph{Context parallelism.}
To train on sequences that are too long for one GPU, context parallelism (CP) splits a sequence of length $L$ across a group of $P$ GPUs, so each GPU holds only $n = L/P$ queries. Ring Attention~\cite{liu2024ringattention} and DeepSpeed-Ulysses~\cite{jacobs2023ulysses}, the widely-used CP methods designed for full attention, do not fit learned sparse attention (\S\ref{sec:overlap}). Existing systems instead reuse the KV all-gather method of Llama~3~\cite{llama3}.
Before selection, the forward pass synchronously all-gathers the indexer keys and the KV states over the CP group. The backward pass reduce-scatters their gradients.

\paragraph{Activation recomputation.}
The backward pass needs the activations of the forward pass. At long context that storage can exceed GPU memory, so training often uses activation recomputation~\cite{chen2016training}. It discards some activations during the forward pass and recomputes them when the backward pass needs them, trading compute for memory. Under full activation recomputation, each layer keeps only its input and reruns its forward pass right before the backward pass. Selective recomputation~\cite{korthikanti2023reducing,pytorchsac} keeps more activations and reruns only parts of the layer, such as the core attention. It works at the granularity of operators, either keeping an operator's output or rerunning the operator in full.

\subsection{The Selection Bottleneck}
\label{sec:inefficiencies}

\begin{figure}[t]
\centering
\includegraphics[width=\columnwidth]{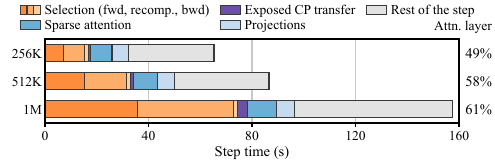}
\caption{One training step of GLM-5.1 in Megatron-LM on 512 H20 GPUs with full recomputation and 8K tokens per GPU, first pipeline stage. Step times exclude the pipeline wait. The attention layer is in color.}
\label{fig:bottleneck}
\end{figure}

Although sparse attention is cheap, the attention layer remains the bottleneck of long-context training, and selection accounts for most of it. Figure~\ref{fig:bottleneck} profiles GLM-5.1 training in Megatron-LM, a state-of-the-art training system, on 512 H20 GPUs with 8K tokens per GPU. Excluding the pipeline wait, the attention layer takes 49\% of the step at 256K tokens and 61\% at 1M. At 1M tokens, selection alone takes 47\% of the step, 6.6$\times$ as long as sparse attention.

\begin{figure}[t]
\centering
\includegraphics[width=\columnwidth]{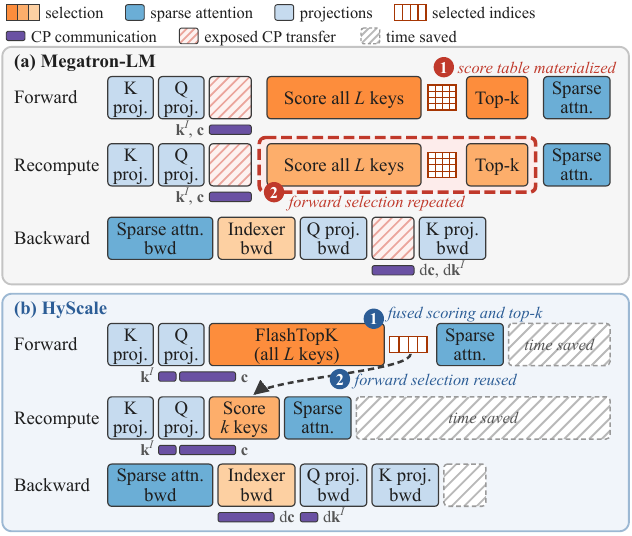}
\caption{One learned sparse attention layer on one GPU under CP and full recomputation (not to scale). Red numbers in (a) mark the score table (\waste{1}) and the repeated selection (\waste{2}). Blue numbers in (b) mark where \flashtopk (\fix{1}) and \sparsereplay (\fix{2}) remove them. \sysname also overlaps the CP collectives with computation. $\mathbf{k}^I$ denotes the indexer keys, and $\mathbf{c}$ the KV states.}
\label{fig:passes}
\end{figure}

Figure~\ref{fig:passes}a shows how Megatron-LM executes one learned sparse attention layer in a training step and marks two inefficiencies in selection (\waste{1} and \waste{2}).

\paragraph{Inefficiency 1: Materialized scores.}
Selection materializes every score before selecting any (\waste{1}). Current systems run it as two kernels. The scoring kernel scores the $n$ local queries against all $L$ keys and writes the scores to GPU memory as an $n \times L$ table. The top-$k$ kernel then reads the table back and keeps $k$ entries per query. In the 1M-token step ($n{=}8$K, $k{=}2048$), the table of one layer takes 32\,GiB on each GPU, whereas the selected indices take only 64\,MiB. Since top-$k$ keeps only 0.2\% of the scores, materializing them wastes GPU memory and memory traffic, a cost that grows with the context.

\paragraph{Inefficiency 2: Repeated selection.}
Under full activation recomputation, selection runs twice in every training step (\waste{2}). Before each layer's backward pass, the rerun of its forward pass scores all $L$ keys and selects the top-$k$ again. As Figure~\ref{fig:bottleneck} shows, this rerun takes about as long as selection in the forward pass. At 1M tokens, it thus accounts for half of the 47\% of the step that selection takes, and like the forward selection, its cost grows with the context.

\subsection{Key Insight}
\label{sec:insight}

Both inefficiencies stem from the same root cause. Current systems handle the output of selection, the selected indices, like a regular activation. They compute it with two dense operators, scoring and top-$k$, so the score table becomes an intermediate tensor in GPU memory (\waste{1}). Recomputation regenerates it by rerunning both operators (\waste{2}).

Our key insight is that \textbf{the output of selection is not a continuous activation, but rather a discrete set of indices}. Two properties of learned sparse attention support this view. First, the score table serves only to find these indices. Sparse attention reads only the selected entries, and the indexer loss reads only their scores, so nothing reads the scores of the other keys. Second, the backward pass needs only these indices and their scores. Top-$k$ is not differentiable, so no gradient flows through selection, and the backward pass never needs the score table or any other intermediate result of selection (Appendix~\ref{app:backward}).

This insight motivates our design in two ways. First, selection need not materialize the score table. Since only the scores of the selected entries are used, the scoring kernel can keep or drop each score as soon as it computes it (\flashtopk, \S\ref{sec:selection}). Second, recomputation need not repeat selection. Since the backward pass needs only the selected indices and their scores, recomputation can reuse the indices from the forward pass and rescore only the selected entries instead of all entries (\sparsereplay, \S\ref{sec:replay}).

\section{\sysname Design}
\label{sec:design}

\subsection{Overview}
\label{sec:overview}

\sysname builds on Megatron-LM and changes only its learned sparse attention layers. Every GPU in the CP group runs one copy of \sysname (Figure~\ref{fig:arch}). Two techniques apply the insight of \S\ref{sec:insight} to remove both inefficiencies (Figure~\ref{fig:passes}b). \flashtopk (\S\ref{sec:selection}) selects each query's top-$k$ while scoring, without writing the score table (\fix{1}). \Sparsereplay (\S\ref{sec:replay}) reuses the forward indices and rescores only their $k$ entries in the recomputation (\fix{2}). Accelerating selection exposes two remaining costs, CP communication and the sparse kernels. \Overlap (\S\ref{sec:overlap}) and locality-aware sparse kernels (\S\ref{sec:kernels}) reduce them respectively.

\subsection{\flashtopk}
\label{sec:selection}

\paragraph{Fused sampling-based top-$k$ selection.}
Current systems run unfused selection with two kernels. The scoring kernel writes all $n \times L$ scores to GPU memory, and a top-$k$ kernel reads them back to keep $k$ per query (\waste{1}). Fused selection merges the two with a sampled threshold. It first estimates the threshold from a sample of keys. The scoring kernel then compares every score with it on chip and writes only the passing keys to a small buffer per query. If every key of the true top-$k$ passes, selecting from these candidates is exact. The score table thus never reaches GPU memory.

The performance of this design depends on the quality of the sample. A poor sample would give a threshold far below the true $k$-th score, which lets many keys through, adds memory traffic and selection work, and can overflow the buffer. LiteTopK~\cite{litetopk} pioneered this design for inference. Its threshold is the $k$-th largest score among the keys that nearby queries selected, as nearby queries tend to select similar keys. This guaranteed lower bound never exceeds the true $k$-th score and lets every key of the true top-$k$ pass. If too many keys still pass, it raises the threshold during the scan so that the candidates fit a fixed-size buffer. DoSA~\cite{dosa} further speeds up this design with key pruning: it first computes a cheap upper bound of each key's score, and skips the full scoring of keys whose bound falls below the threshold. Both designs, however, build their sampling and pruning around inference prefill and break down in three ways during training (Table~\ref{tab:design}).

\paragraph{Challenges in training.}
\emph{C1: Neighbor sampling breaks under CP.} LiteTopK and DoSA use \emph{neighbor sampling}: they sample the keys that nearby queries selected. This sample covers most of the true top-$k$, keeping their guaranteed lower bound tight. Inference has these selections at hand, as it processes a long prompt chunk by chunk on one GPU~\cite{agrawal2024sarathi}. Under CP, however, other GPUs score nearby queries at the same time (\S\ref{sec:bg-training}). Their selections are thus not yet available. Without them, a guaranteed lower bound must come from some other sample of the query's context. Such a bound can be so loose that far too many keys pass, slowing down selection.

\begin{table}[t]
\centering
\fontsize{8}{9}\selectfont
\setlength{\tabcolsep}{3.5pt}
\caption{Design of fused sampling-based top-$k$ selection.}
\label{tab:design}
\begin{tabular}{@{}lccc@{}}
\toprule
 & LiteTopK & DoSA & \textbf{\flashtopk} \\
\midrule
Sample from & Neighbors & Neighbors & \textbf{Own context} \\
Threshold & Guaranteed & Guaranteed & \textbf{Optimistic} \\
Works under CP & \xmark & \xmark & \textbf{\cmark} \\
Key pruning & \xmark & Always & \textbf{Adaptive} \\
\bottomrule
\end{tabular}
\end{table}

\begin{figure}[t]
\centering
\includegraphics[width=\columnwidth]{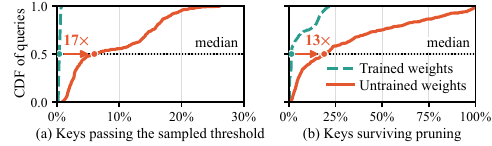}
\caption{Sampling and pruning efficiency of DoSA (GLM-5.1, 1M). (b) uses the exact $k$-th score as threshold.}
\label{fig:c2}
\end{figure}

\emph{C2: Neighbor sampling and key pruning fail in early training.} They both rely on a trained indexer, but training starts from untrained weights. First, with random weights, nearby queries no longer select similar keys. Neighbor sampling thus sets the threshold far too low: 17$\times$ more keys pass at the median than with trained weights (Figure~\ref{fig:c2}a). Second, key pruning skips far fewer keys: even with the exact $k$-th score as threshold, 13$\times$ more keys survive at the median (Figure~\ref{fig:c2}b). In our practice, as a result, DoSA can run even slower than unfused selection with untrained weights.

\emph{C3: Training requires deterministic selection.} Training needs bitwise reproducible kernels to replay a run and debug loss spikes or hardware faults~\cite{dsv4,mok}. Existing sampling-based operators break this, as they append the passing keys to each query's buffer with atomics. The key order thus varies across runs and changes the output in two ways. First, when several keys tie at the $k$-th score, the order decides which are kept. Second, it decides the floating-point rounding of the sum in sparse attention. Sorting each query's candidates by position fixes both, but costs even more than selecting the top-$k$ itself.

\paragraph{\flashtopk.}
To bridge the gap between sampling-based top-$k$ selection and training, we propose \flashtopk{} (Table~\ref{tab:design}). Its key insight is that \emph{the threshold need not be a guaranteed lower bound, as long as it rarely fails and each failure is cheap to recover from}. The threshold then only has to be close to the $k$-th score, yet selection stays exact. \flashtopk{} therefore estimates the threshold from each query's own context (C1, C2). It also prunes keys only when a sample predicts a gain (C2), and makes its selection deterministic (C3). Appendix~\ref{app:flashtopk} details its kernels.

\paragraph{Optimistic strided sampling (C1, C2).}
\flashtopk gives up the guarantee and estimates each query's threshold from a strided sample of its own context. It thus needs no other query's selection and works under CP. First, \flashtopk samples once for all queries on a GPU: it takes the middle key of each of $M = \min(\max(2048, L/128), 8192)$ equal segments of the $L$ keys. Second, \flashtopk scores this sample for all queries in one matrix multiplication. Query $t$ uses the $m_t$ samples within its context of $L_t$ keys, each standing for about $L_t/m_t$ keys. Third, \flashtopk sets the threshold to the $\lceil 2.5km_t/L_t \rceil$-th best of them, an estimate of the $2.5k$-th best score. About $2.5k$ candidates thus pass, $2.5\times$ the $k$ keys to select. The sample spans the whole context, so the threshold rarely rises above the $k$-th best score. For the few queries that still end with fewer than $k$ candidates or overflow the buffer, the GPU redoes their selection over all keys without host synchronization.

\paragraph{Adaptive key pruning (C2).}
\flashtopk uses DoSA's score bound but prunes adaptively, unlike DoSA's always-on pruning. In early training, the bound loosens and pruning may not pay off (C2). \flashtopk thus decides before each selection whether to prune. It first estimates the share of keys the bound keeps, on one query in 16 against 256 sampled keys. If this share is at most 20\%, a pruning pass marks keys that any of 8 adjacent queries could select, and the scan scores only these keys. Otherwise, it scans every key.

\paragraph{Deterministic selection (C3).}
\flashtopk keeps the fast atomic writes in its kernels and restores determinism afterwards. It orders keys only within the smallest sets whose order affects the output. First, when more keys share the $k$-th highest score than there are places left in the selection, the select kernel runs a radix select~\cite{alabi2012kselection} over the positions of just these tied keys and keeps those with the smallest positions. The selected set then depends only on the scores. Second, a radix sort~\cite{merrill2011radix} orders the $k$ selected entries of each query by score, breaking ties by position, so sparse attention sums over them in the same order in every run. Two runs on the same inputs thus produce the same exact top-$k$, bit for bit.

\subsection{Selection Replay}
\label{sec:replay}

Under activation recomputation, current systems rerun selection before the backward pass (\waste{2} in Figure~\ref{fig:passes}a). They score all $L$ keys of every query and select the top-$k$ again, although the layer input has not changed and the selection is the same, which doubles the cost of selection.

Our insight is that this rerun is unnecessary, because the output of selection is only indices. Top-$k$ is not differentiable, so gradients reach the $k$ selected entries but never pass through the selection itself. The backward pass thus needs only the selected indices and the $k$ scores at them (\S\ref{sec:bg-lsa}), not the scoring of all $L$ keys. This insight motivates us to keep the indices from the forward pass and, in the recomputation, rescore only the entries they point to, cutting selection recomputation from $O(L)$ to $O(k)$ per query.

\begin{figure}[t]
\centering
\includegraphics[width=\columnwidth]{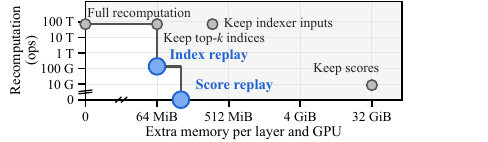}
\caption{Costs of selection recomputation policies for one GLM-5.1 layer at 1M tokens on one GPU. Appendix~\ref{app:replay-costs} derives them.}
\label{fig:replay-pareto}
\end{figure}

The challenge is that existing selective recomputation~\cite{korthikanti2023reducing,pytorchsac} cannot rescore only these keys. It decides for each operator whether to keep the operator's output or to rerun the operator in full. Even if it keeps the indices that top-$k$ outputs, it cannot rerun the scoring operator for only the $k$ entries they point to. Every policy it can express therefore wastes either memory or computation (Figure~\ref{fig:replay-pareto}). Keeping all activations or the score table avoids rescoring but wastes memory, because both grow with the context. Keeping nothing, the indexer's inputs, or only the top-$k$ indices wastes computation, because the recomputation still scores all $L$ keys to obtain the $k$ scores that the backward pass needs.

\paragraph{Selection replay.}
To address this challenge, we design \emph{selection replay}, a recomputation policy for selection that decides what to keep and what to rerun per entry rather than per operator. The forward pass keeps the selection of each query, and the recomputation computes at most the $k$ scores that the backward pass reads. Replay comes in two variants that differ in what the forward pass keeps. \emph{Index replay} keeps only the $k$ indices of each query and rescores only the $k$ entries they point to. \emph{Score replay} also keeps the $k$ scores and rescores nothing.

\paragraph{Cost comparison.}
Table~\ref{tab:replay-costs} in Appendix~\ref{app:replay-costs} derives the cost of every policy, and Figure~\ref{fig:replay-pareto} plots these costs for one GLM-5.1 layer at 1M tokens with 8K tokens per GPU. Both variants of replay sit in the lower-left corner, while every operator-level policy in the figure costs at least two orders of magnitude more in memory or in recomputation. We use \indexreplay by default, since its rescoring costs under 1\% of a training step at 1M and it saves about 5\,GiB per GPU over \scorereplay in GLM-5.1.

\paragraph{Implementation of index replay.}
We describe \emph{index replay} in detail. Figure~\ref{fig:passes}b shows \indexreplay in one layer of \sysname (\fix{2}). First, the forward pass scores all $L$ keys and selects the top-$k$ as in \S\ref{sec:selection}, and keeps the $k$ selected indices of each query. Second, the recomputation does not score all keys or select again. It loads the kept indices, scores only the $k$ entries they point to with a gather-and-score kernel, and runs sparse attention on the same indices. Third, the backward pass runs exactly as without replay.

To save GPU memory, the kept indices can be offloaded to pinned host memory with nearly zero overhead. The forward pass copies them to the host on a separate stream right after selection, overlapping with sparse attention. The recomputation copies them back when it enters the attention layer, overlapping with the projections.
\paragraph{Generality.} \Sparsereplay applies to any learned sparse attention that selects by top-$k$, whether its entries are tokens as in DSA or blocks as in CSA and QSA. Each such layer learns its selection only through the $k$ selected scores (\S\ref{sec:bg-lsa}), so replay rescores only these entries (\S\ref{sec:eval-general}). Replay is also compatible with existing recomputation policies. It keeps the interface of the operator that scores and selects, so policies that decide per module what to rerun, such as Megatron-LM's, work with replay unchanged.

\subsection{Holistic System Implementation}
\label{sec:comm}

The two techniques cut the work of selection but leave two other costs of learned sparse attention. First, CP communication runs synchronously before selection and sparse attention. Second, sparse kernels incur heavy global memory traffic. \sysname addresses each cost with a property of learned sparse attention. Selection and sparse attention consume different CP data. \Overlap therefore hides each transfer behind computation independent of it (\S\ref{sec:overlap}). Adjacent queries also select largely the same keys. Locality-aware sparse kernels thus process two adjacent queries in one thread block (\S\ref{sec:kernels}).

\subsubsection{Decoupled CP Overlap for Selection and Attention}
\label{sec:overlap}
\paragraph{Bottleneck of CP communication.}
Under CP, each GPU holds only its own queries but needs the indexer keys and the KV states of all earlier tokens. Existing systems therefore all-gather both before selection, in the forward pass and again in the recomputation, and reduce-scatter their gradients in the backward pass (Figure~\ref{fig:passes}a). These collectives are large and synchronous. Figure~\ref{fig:bottleneck} shows the resulting cost when Megatron-LM trains GLM-5.1: the exposed CP transfer grows 4.8$\times$ from 256K tokens (CP32) to 1M (CP128).

\paragraph{Limits of existing CP optimizations.}
Existing CP optimizations designed for full attention do not carry over to learned sparse attention. Ring Attention~\cite{liu2024ringattention} overlaps each transfer with the attention that consumes it, but sparse attention reads only a few selected entries per query and has too little computation to hide the transfer. DeepSpeed-Ulysses~\cite{jacobs2023ulysses} saves communication by splitting attention across GPUs by heads, so that each GPU gathers only the KV of its own heads. In DSA, however, all attention heads read the same KV latent and all indexer heads read the same indexer key, so each GPU would still gather the whole latent and key of every token, exactly what all-gather transfers.

\paragraph{Opportunity from disjoint inputs.}
To optimize transfers in learned sparse attention, we propose \overlap. Its key insight is that the two steps of learned sparse attention consume different CP data: selection reads only the indexer keys, and sparse attention, which runs after selection, reads only the KV states.

\paragraph{Design.}
\sysname decouples the CP communication of the two steps, sending the indexer keys and KV states separately, and overlaps each transfer with computation that does not consume it (Figure~\ref{fig:passes}b). Each transfer is issued as soon as its data is ready and waited for only before its first consumer. In the forward pass and recomputation, the indexer-key gather overlaps the query projections, and the KV-state gather overlaps the query projections and selection. In the backward pass, the reduce-scatter of the KV-state gradient overlaps the indexer backward, and that of the indexer-key gradient overlaps the query projections' backward.
\subsubsection{Locality-Aware Sparse Kernels}
\label{sec:kernels}
\begin{figure}[t]
\centering
\includegraphics[width=\columnwidth]{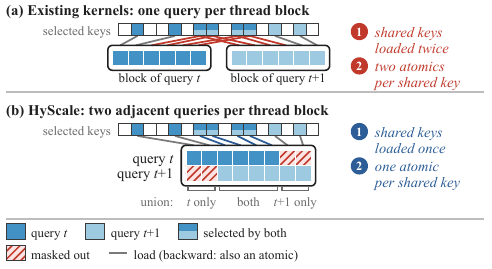}
\caption{Locality-aware sparse kernels. Existing kernels (a) assign each thread block a single query; \sysname (b) processes two adjacent queries in one thread block over the union of their selections. Each row in a block marks the keys that one query uses.}
\label{fig:kernels}
\end{figure}
\paragraph{Bottleneck of sparse kernels.}
Besides selection, learned sparse attention adds three sparse kernels to training: the sparse attention forward, its backward, and the indexer backward (Figure~\ref{fig:passes}). Existing implementations~\cite{flashmla,cudnndsa,tilelang} assign each thread block a single query (Figure~\ref{fig:kernels}a). This per-query split costs L2 traffic in two forms. First, \emph{per-query loads} (\waste{1}): the forward kernel loads each query's selected entries separately. Second, \emph{per-query atomics} (\waste{2}): the backward kernels load entries the same way. They also add each query's gradient to its entries with separate atomics. Profiling the backward kernels for GLM-5.1 at 512K tokens on H800 confirms both costs. The L2 cache runs at 53--65\% of its peak throughput, and warps stall on memory in 40--44\% of their cycles. The tensor cores are only 21--32\% busy.

\paragraph{Opportunity from selection locality.}
To address these bottlenecks, we propose locality-aware sparse kernels. Their key insight is \emph{selection locality}: adjacent queries select largely the same keys, as prior work has observed~\cite{litetopk,pivot,dosa}. In GLM-5.1, for example, two adjacent queries share 71\% of their \mbox{top-2048} keys on average, measured over 21 of its 78 layers on real documents of 128K and 192K tokens; the union of their selections is thus only 1.29 times one selection. Locality-aware kernels therefore process two adjacent queries in one thread block over the union of their selections (Figure~\ref{fig:kernels}b), which reduces both bottlenecks. First, each kernel loads a shared key once for both queries (\fix{1}). Second, the backward kernels sum both queries' gradients of a shared key before adding them with one atomic (\fix{2}).

\paragraph{Design.}
\leavevmode Once per layer, a lightweight preprocessing step builds the unions of all adjacent query pairs in one batch. All three sparse kernels reuse these unions. Each thread block reads the union of its own pair. The union lists the keys selected only by the first query, then the shared keys, then the keys selected only by the second (Figure~\ref{fig:kernels}b). The first query uses the first two parts. The second query uses the last two. Each query masks out the remaining part, the hatched cells, so outputs and gradients stay exact. This design spends otherwise idle tensor-core cycles on the masked part to save L2 traffic on the shared keys.

\subsubsection{Implementation Details}
\label{sec:impl}
\sysname is implemented on top of Megatron-LM with about 3.4K lines of Python and 14.7K lines of GPU kernels (CUDA, Triton, CuTe DSL). Besides the two optimizations above, four techniques target two overheads in Megatron-LM's attention layers: host synchronizations and wasted GPU work. The first two remove host synchronizations. (1)~\emph{CP layout cache}: Megatron-LM rebuilds the CP layout with its masks in every layer and pass, synchronizing with the host for every CP rank. \sysname builds it once per micro-batch. (2)~\emph{Sequence boundaries cache}: Megatron-LM's RoPE also synchronizes in every call to read the sequence boundaries, which \sysname copies to the host once per micro-batch. The other two remove wasted GPU work. (3)~\emph{Kernel fusion}: \sysname fuses chains of elementwise operators, such as the indexer's KL loss, normalizations, and RoPE. (4)~\emph{Permutation-free KV gather}: Megatron-LM reorders the gathered KV states into token order in every pass, so that sparse attention can index them by token. \sysname instead adjusts the indices to the gathered order.

\section{Evaluation}
\label{sec:eval}
\subsection{Setup}
\label{sec:eval-setup}
\paragraph{Testbed.} Our experiments run on an H20 cluster and an H800 cluster. Each H20 node has eight NVIDIA H20 GPUs with 96\,GB of memory, two 96-core AMD EPYC 9K84 CPUs, and 2.2\,TiB of DRAM. Each H800 node has eight NVIDIA H800 GPUs with 80\,GB of memory, two 48-core Intel Xeon Platinum 8476C CPUs, and 2\,TiB of DRAM. Nodes in both clusters connect through eight 400\,Gb/s RoCE NICs (NVIDIA ConnectX-7).

\paragraph{Baseline and configuration.} We compare \sysname with Megatron-LM, which runs Megatron Core 0.19.2 with the cuDNN DSA backend. \sysname builds on the same code base, ensuring the two systems differ only in the attention layers. All runs use BF16, zigzag all-gather CP, and full activation recomputation unless noted otherwise. In \sysname, \indexreplay and the offloading of its indices are enabled.

\subsection{End-to-End Training}
\label{sec:eval-e2e}
We train the full GLM-5.1 from its released weights on 512 H20 GPUs, fixing the global batch at 4M tokens and the tokens per GPU at 8K. Each sample is a single sequence of 256K, 512K, or 1M tokens from real documents. The runs use PP8$\times$CP32$\times$DP2$\times$EP64 at 256K, PP8$\times$CP64$\times$EP64 at 512K, and PP4$\times$CP128$\times$EP128 at 1M.

\begin{figure}[t]
\centering
\includegraphics[width=\columnwidth]{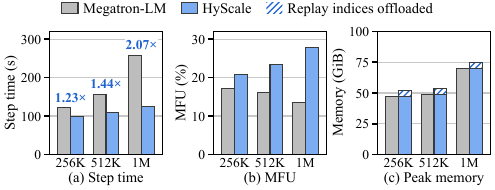}
\caption{Full GLM-5.1 on 512 H20 GPUs. Both systems force balanced MoE routing and use momentum-free SGD with Megatron-LM's distributed optimizer. Step times average steps 4 to 10.}
\label{fig:eval-e2e}
\end{figure}

\paragraph{\sysname trains much faster with similar memory.} \sysname is up to 2.07$\times$ faster than Megatron-LM, and the speedup grows with the context, from 1.23$\times$ at 256K to 2.07$\times$ at 1M (Figure~\ref{fig:eval-e2e}a). \sysname achieves this speedup with about the same peak GPU memory (Figure~\ref{fig:eval-e2e}c). Even without offloading the replay indices (\S\ref{sec:replay}), it would only use at most 11\% more.

\begin{figure}[t]
\centering
\includegraphics[width=\columnwidth]{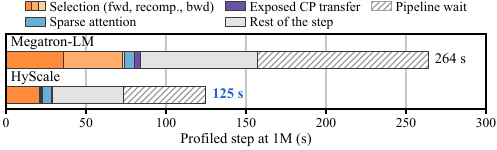}
\caption{Breakdown of one full GLM-5.1 training step at 1M.}
\label{fig:e2e-breakdown}
\end{figure}

\begin{figure}[t]
\centering
\includegraphics[width=\columnwidth]{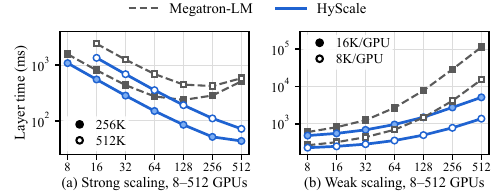}
\caption{(a) Strong and (b) weak CP scaling of one GLM-5.1 attention layer on H20. Layer time covers the forward, recomputation, and backward passes on the slowest rank.}
\label{fig:eval-scaling}
\end{figure}

\paragraph{\sysname becomes more efficient at longer contexts.} From 256K to 1M, with the same 4M tokens in every step, \sysname's step time grows only 1.27$\times$, and its MFU even rises, from 20.9\% to 27.8\% (Figure~\ref{fig:eval-e2e}b). Over the same range, Megatron-LM's step time grows 2.13$\times$, and its MFU falls from 17.1\% to 13.5\%.

\sysname's advantage stems mainly from faster selection, the main bottleneck that grows with the context. On the first pipeline stage at 1M (Figure~\ref{fig:e2e-breakdown}), \flashtopk cuts the forward selection from 35.9\,s to 20.8\,s, and \sparsereplay cuts the recomputed selection from 37.0\,s to 1.0\,s. Faster attention layers also halve the pipeline wait, from 106.6\,s to 51.7\,s.

\subsection{Context-Parallel Scalability}
\label{sec:eval-scaling}
We benchmark a single GLM-5.1 attention layer on 8 to 512 H20 GPUs, with the CP size equal to the GPU count, under two standard settings~\cite{jacobs2023ulysses} (Figure~\ref{fig:eval-scaling}). First, strong scaling fixes the context length and adds GPUs. Second, weak scaling fixes the queries per GPU and grows the context with the GPUs.

\paragraph{\sysname keeps speeding up with more GPUs.} At both 256K and 512K context, \sysname runs faster every time the GPU count doubles (Figure~\ref{fig:eval-scaling}a). Up to 256 GPUs, each doubling cuts the layer time by 1.65 to 1.96$\times$. Beyond 256 GPUs, the gain drops to 1.18 to 1.54$\times$, because each GPU holds too few queries to hide the CP transfers. \sysname's speedup over Megatron-LM grows with the GPU count, from 1.44 to 11.7$\times$ at 256K and from 1.79 to 8.1$\times$ at 512K. In fact, Megatron-LM stops improving as early as 128 GPUs, and even slows down beyond 128 GPUs. This slowdown mainly stems from rebuilding the CP layout in every layer, incurring host synchronizations that grow linearly with the GPU count. \sysname avoids this cost with its CP layout cache (\S\ref{sec:impl}). The cache alone does not explain \sysname's speedup, however. At 1M context on 128 GPUs, all system optimizations together, the cache included, save only 11\% of Megatron-LM's layer time (\S\ref{sec:eval-breakdown}). Even if the cache saved all of it, \sysname would still run 2.6$\times$ faster than Megatron-LM.

\paragraph{\sysname's CP extends the context gracefully.} \sysname is faster than Megatron-LM at all context lengths, and its layer time grows far more slowly as the context grows (Figure~\ref{fig:eval-scaling}b). With 16K queries per GPU, from 128K to 8M context, \sysname's layer time grows only 11$\times$, while Megatron-LM's grows 190$\times$. \sysname is 1.24$\times$ faster than Megatron-LM at 128K context, and by 8M, this speedup climbs to 22.3$\times$. The runs with 8K queries per GPU show the same trend, with an 11.2$\times$ speedup at 4M.

\subsection{Ablation Study}
\label{sec:eval-breakdown}
\begin{figure}[t]
\centering
\includegraphics[width=\columnwidth]{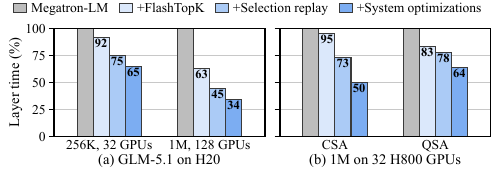}
\caption{Cumulative ablation of one attention layer, as a share of Megatron-LM's layer time: (a) GLM-5.1 on H20 at 8K queries per GPU, (b) CSA and QSA on 32 H800 GPUs.}
\label{fig:eval-ablation}
\end{figure}
We measure the layer time of one GLM-5.1 attention layer at 256K context on 32 GPUs and at 1M on 128 GPUs, with the CP size equal to the GPU count and 8K queries per GPU (Figure~\ref{fig:eval-ablation}a). Starting from Megatron-LM, we add \flashtopk (\S\ref{sec:selection}), \sparsereplay (\S\ref{sec:replay}), and the system optimizations of \S\ref{sec:comm} one at a time. Each step is measured on top of the previous ones. We report each step's saving as a share of Megatron-LM's layer time.

\paragraph{\flashtopk and \sparsereplay bring most of the speedup.} \looseness=-1 At 1M context, \sysname saves 66\% of Megatron-LM's layer time (Figure~\ref{fig:eval-ablation}a). The two techniques deliver 84\% of this saving. \flashtopk saves 37\% of the layer time, because it removes the score table from both the forward pass and the recomputation. \Sparsereplay saves another 18\%, because it skips the recomputed selection. The system optimizations save the remaining 11\%. \flashtopk's share also grows fastest, from 8\% at 256K to 37\% at 1M, because it replaces Megatron-LM's unfused selection, whose cost grows faster than the context. Megatron-LM cannot hold the 32\,GiB score table at 1M, so it splits the queries into chunks whose score tables fit a fixed memory limit. Longer contexts leave fewer queries per chunk, too few at 1M to keep the GPU busy. \flashtopk avoids this trade-off: it writes no score table, so it scores all queries at once. \Sparsereplay keeps 16--18\% at both contexts, because \flashtopk already accelerates the selection it skips.

\subsection{Generality}
\label{sec:eval-general}
This experiment tests whether \sysname carries over to block-level architectures. We implement the CSA layer of DeepSeek-V4-Flash~\cite{dsv4} and the QSA layer of Qwen3.8-Flash-Next~\cite{qwen38next} in Megatron-LM from their papers. We repeat \S\ref{sec:eval-breakdown}'s ablation on 32 H800 GPUs at 1M context with CP32 (Figure~\ref{fig:eval-ablation}b).

\paragraph{\sysname generalizes across learned sparse attention.} \sysname runs CSA 2.01$\times$ and QSA 1.57$\times$ faster than Megatron-LM. Every ablation step speeds up both layers, showing that \sysname can also accelerate block-level architectures. \flashtopk's gain on block-level layers is smaller than on token-level DSA (\S\ref{sec:eval-breakdown}), since they score 4$\times$ fewer keys per query, making selection less significant.

\subsection{\flashtopk Efficiency}
\label{sec:eval-flashtopk}
\begin{figure}[t]
\centering
\includegraphics[width=\columnwidth]{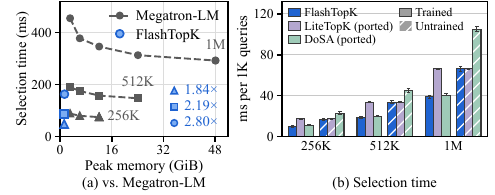}
\caption{\flashtopk efficiency for GLM-5.1 on one H20 GPU.}
\label{fig:eval-flashtopk-new}
\end{figure}
\begin{figure}[t]
\centering
\includegraphics[width=\columnwidth]{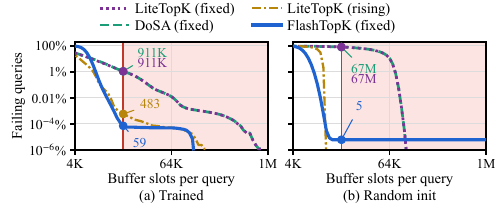}
\caption{Sampling efficiency: share of failed queries at each buffer size. Labels count failures at \flashtopk's default 16K-slot candidate buffer (red line), 1\,GiB per GPU.}
\label{fig:eval-flashtopk}
\end{figure}
\paragraph{Operator efficiency.} We compare \flashtopk with Megatron-LM's cuDNN selection on one H20 GPU. We benchmark all selections captured from full GLM-5.1 training at 256K, 512K, and 1M context, on all 78 layers and CP ranks. Each selection picks $k=2048$ keys for 8K queries. Among fused sampling-based methods, only \flashtopk supports CP and determinism (\S\ref{sec:selection}). Moreover, LiteTopK and DoSA release kernels only for FP8 on Blackwell GPUs. We thus reimplement both in BF16 on H20 following their code, and run them without CP. Both are also non-deterministic, while \flashtopk keeps its deterministic sort.

\paragraph{\flashtopk removes the memory--time trade-off of selection.} \flashtopk uses the least memory and runs the fastest at every context (Figure~\ref{fig:eval-flashtopk-new}a). Including its buffers, it peaks at only 1.3--1.4\,GiB per layer. Megatron-LM must instead trade memory for time. With its default 1\,GiB limit on the score table, it peaks at 3.1\,GiB but runs 1.84--2.80$\times$ slower than \flashtopk. With the whole score table, it needs 12--48\,GiB and still runs 55--80\% slower. The limit saves memory, but it splits the queries into chunks too small to keep the GPU busy (\S\ref{sec:eval-breakdown}). \flashtopk avoids both costs because it never writes the score table.

\paragraph{\flashtopk is fast with both trained and untrained weights.} It matches or outperforms LiteTopK and DoSA in both settings (Figure~\ref{fig:eval-flashtopk-new}b). With trained weights, \flashtopk runs \mbox{1.70--1.79$\times$} faster than LiteTopK and \mbox{1.04--1.10$\times$} faster than DoSA. \flashtopk and DoSA both outpace LiteTopK mainly through pruning. Pruning cuts \flashtopk's time by \mbox{39--42\%}, while LiteTopK never prunes. \flashtopk further outpaces DoSA because neighbor sampling forces DoSA's query blocks to run one after another. With untrained weights, \flashtopk runs \mbox{1.34--1.58$\times$} faster than DoSA and \mbox{0.99--1.06$\times$} as fast as LiteTopK. The lead over DoSA is large because DoSA always prunes, and pruning adds large overhead once the bound loosens (C2 in \S\ref{sec:selection}). With neighbor sampling and a fixed threshold, 20\% of DoSA's queries also overflow at 1M. \flashtopk stays on par with LiteTopK because pruning rarely pays off, yet adaptive pruning still turns on in some calls. Overall, optimistic strided sampling and adaptive pruning keep \flashtopk fast and robust.

\paragraph{Sampling efficiency.} Next, we compare \flashtopk's optimistic strided sampling with the neighbor sampling of LiteTopK and DoSA. We count the candidates of all 81.8M queries in one GLM-5.1 training step at 1M, with the trained weights and with a randomly initialized indexer. LiteTopK raises its threshold during the scan (\S\ref{sec:selection}), while DoSA and \flashtopk keep theirs fixed. We show LiteTopK both ways.

\paragraph{Optimistic thresholds outperform guaranteed lower bounds.} \leavevmode With fixed thresholds and trained weights, \flashtopk fails on over 15,000$\times$ fewer queries than LiteTopK and DoSA at its 16K-slot candidate buffer (Figure~\ref{fig:eval-flashtopk}a). With random weights, both baselines fail on 82\% of the queries, and \flashtopk on only 5 of the 81.8M (Figure~\ref{fig:eval-flashtopk}b). Raising the threshold during the scan cuts LiteTopK's failures to 483 with trained weights, still 8$\times$ more than \flashtopk, and to none with random weights. \flashtopk's lightweight recovery rescores only the failed queries on the GPU, with near-zero overhead.

These results reflect the limits of neighbor sampling in training. Its bound comes from the keys that nearby queries selected. Their $k$-th best can thus lie deep in this query's ranking and let many extra keys pass. With random weights, nearby queries also barely share keys (C2 in \S\ref{sec:selection}). Under CP, their selections are not even available (C1). \flashtopk instead samples evenly from each query's own context. Combined with lightweight recovery, optimistic strided sampling thus suits training better.

\section{Related Work}
\label{sec:related}
\paragraph{Selection acceleration for learned sparse attention.} HPC-Ops Top-K~\cite{hpcopstopk}, GVR~\cite{gvr}, DeepSelect~\cite{deepselect}, and Prof-K~\cite{profk} accelerate the top-$k$ step but leave scoring in a separate kernel. The score table thus still makes a round trip through GPU memory. Fusing scoring with top-$k$ selection removes it. LiteTopK~\cite{litetopk} pioneered this fusion with sampling-based selection, taking its threshold from nearby queries' selections. DoSA~\cite{dosa} further skips part of the scoring with cheap per-key bounds. Both, however, build their sampling around prefill and face challenges in training (\S\ref{sec:selection}). \flashtopk addresses them with optimistic strided sampling, adaptive key pruning, and deterministic selection. Orthogonally, PIVOT~\cite{pivot}, IndexCache~\cite{indexcache}, HISA~\cite{hisa}, and MISA~\cite{misa} cut selection work by reusing selections or scoring fewer entries or heads.

\paragraph{Activation recomputation.} Activation recomputation saves memory by rerunning the forward pass during backpropagation~\cite{chen2016training}, at the cost of extra computation. Prior work cuts this cost by choosing what to recompute. Checkmate~\cite{jain2020checkmate} and DTR~\cite{kirisame2021dtr} pick tensors offline or at runtime; selective recomputation~\cite{korthikanti2023reducing,pytorchsac} reruns only cheap operators with large outputs. DeepSeek-V4~\cite{dsv4} further makes this choice per tensor by extending automatic differentiation. All of them, however, rerun whole operators. In learned sparse attention, this wastes memory or computation (\S\ref{sec:replay}). \Sparsereplay avoids both by rescoring only the selected entries.

\paragraph{Context parallelism.} The basic CP design all-gathers all KV states before attention~\cite{llama3}, leaving this communication exposed. Two schemes optimize it. Ring Attention~\cite{liu2024ringattention}, Striped Attention~\cite{brandon2023striped}, and DistFlashAttn~\cite{li2024distflashattn} hide it behind attention computation by passing KV blocks around a ring. DeepSpeed-Ulysses~\cite{jacobs2023ulysses} reduces it by splitting attention heads across GPUs. USP~\cite{fang2024usp} and LoongTrain~\cite{gu2024loongtrain} further combine both for hierarchical GPU networks. Both schemes, however, fall short in learned sparse attention (\S\ref{sec:overlap}). Ring designs lack the computation to hide transfers~\cite{bu2026longcontext}, and Ulysses cannot split KV states shared across heads, such as in DSA. \sysname instead hides the KV communication behind selection. Orthogonally, MagiAttention~\cite{magiattention}, ByteScale~\cite{ge2025bytescale}, WLB-LLM~\cite{wang2025wlbllm}, and FlexSP~\cite{wang2025flexsp} balance the CP workload across GPUs.

\section{Conclusion}
\label{sec:conclusion}
We presented \sysname, a long-context training system for learned sparse attention. Its key insight is that selection outputs only indices, not activations. \flashtopk thus selects the top-$k$ without writing the score table, and \sparsereplay skips repeated selection in recomputation. \sysname trains GLM-5.1 on 1M-token sequences 2.07$\times$ faster than Megatron-LM on 512 H20 GPUs and generalizes across learned sparse attention architectures. Future work includes extending \flashtopk to inference, other precisions such as FP8, and newer GPUs such as Blackwell.

\section*{Acknowledgements}
We thank Tian Zheng from NVIDIA and Yeqi Huang from the University of Edinburgh for insightful discussions and helpful feedback.

\bibliography{references}
\bibliographystyle{mlsys2025}

\appendix

\makeatletter\setlength{\@fptop}{0pt}\makeatother
\newcommand{\fn}[1]{\mathsf{#1}}
\newcommand{\bI}{\mathbf{I}}

\section{Formulation of Learned Sparse Attention}
\label{app:formulation}

Consider a sequence of $L$ tokens. For the query at position $t$, let $\mathcal{U}_t$ be the set of entries it can select from. \emph{Selection} scores every entry and keeps the $k$ highest,
\begin{equation}
\mathbf{I}_t = \big(\mathrm{Score}(\mathbf{q}^{I}_{t}, \mathbf{k}^{I}_{u})\big)_{u\in\mathcal{U}_t},
\quad
\mathcal{T}_t = \mathrm{Top\text{-}k}(\mathbf{I}_t),
\label{eq:index}
\end{equation}
where $\mathbf{q}^{I}_{t}$ is the indexer's query, $\mathbf{k}^{I}_{u}$ is the indexer's key of entry $u$, $\mathbf{I}_t \in \mathbb{R}^{|\mathcal{U}_t|}$ is the score vector, and $\mathcal{T}_t \subseteq \mathcal{U}_t$ is the index set of size $k$. \emph{Sparse attention} then attends from query $\mathbf{q}_t$ to the selected entries alone,
\begin{equation}
\mathbf{o}_t = \mathrm{Attn}\big(\mathbf{q}_t, \{\mathbf{c}_u\}_{u\in\mathcal{T}_t}\big),
\label{eq:attn}
\end{equation}
where $\mathbf{c}_u$ is what sparse attention reads for entry $u$, e.g., a KV latent, and $\mathbf{o}_t$ is the output. The indexer learns from a loss of its own,
\begin{equation}
\mathcal{L}^{I} = \lambda \sum_{t} \mathrm{KL}\big(\mathbf{p}_t \,\big\|\, \mathrm{Softmax}(\mathbf{I}_{t,\mathcal{T}_t})\big),
\label{eq:indexer-loss}
\end{equation}
where $\lambda$ is a scalar weight, $\mathbf{p}_t \in \mathbb{R}^{k}$ is the attention distribution of query $t$ over $\mathcal{T}_t$ averaged over heads, and $\mathbf{I}_{t,\mathcal{T}_t} \in \mathbb{R}^{k}$ holds the indexer's scores on those entries. Both $\mathbf{p}_t$ and the indexer's inputs are detached. This loss thus updates only the indexer, through the $k$ selected scores alone.

DSA scores each token with a lightning indexer of $H_I$ heads of dimension $d_I$,
\begin{equation}
\mathrm{Score}(\mathbf{q}^{I}_{t}, \mathbf{k}^{I}_{u}) = \sum_{h=1}^{H_I} w_{t,h}\,\mathrm{ReLU}\big(\mathbf{q}^{I\top}_{t,h}\,\mathbf{k}^{I}_{u}\big),
\label{eq:dsa}
\end{equation}
where $\mathbf{q}^{I}_{t,h} \in \mathbb{R}^{d_I}$ is the query of head $h$, the heads share one key $\mathbf{k}^{I}_{u} \in \mathbb{R}^{d_I}$ per token, and $w_{t,h}$ weighs each head. In MLA, all heads share one compressed KV latent per token. This latent serves as $\mathbf{c}_u$.

\section{Backward Pass of Learned Sparse Attention}
\label{app:backward}

This appendix derives the backward pass of one learned sparse attention layer in the notation of Appendix~\ref{app:formulation}, for DSA with MLA. The backward passes of selection and sparse attention use only the selected entries.

\paragraph{Selection.}
With $z_{t,h,s} = \mathbf{q}^{I\top}_{t,h}\,\mathbf{k}^{I}_{s}$, Equation~\eqref{eq:dsa} reads $I_{t,s} = \sum_{h} w_{t,h}\,\mathrm{ReLU}(z_{t,h,s})$. Let $\hat{\mathbf{p}}_t = \mathrm{Softmax}(\mathbf{I}_{t,\mathcal{T}_t})$ be the indexer's distribution over the selection. The target $\mathbf{p}_t$ is detached, and $\mathcal{T}_t$ is piecewise constant in the scores. The indexer loss of Equation~\eqref{eq:indexer-loss} thus has the score gradient
\begin{equation}
g_{t,s} = \frac{\partial \mathcal{L}^{I}}{\partial I_{t,s}} =
\begin{cases}
\lambda\,(\hat{p}_{t,s} - p_{t,s}) & s \in \mathcal{T}_t,\\
0 & \text{otherwise}.
\end{cases}
\end{equation}
Let $\delta_{t,h,s} = g_{t,s}\,w_{t,h}\,\mathbf{1}[z_{t,h,s} > 0]$. The chain rule gives
\begin{align}
\frac{\partial \mathcal{L}^{I}}{\partial w_{t,h}} &= \sum_{s\in\mathcal{T}_t} g_{t,s}\,\mathrm{ReLU}(z_{t,h,s}), \\
\frac{\partial \mathcal{L}^{I}}{\partial \mathbf{q}^{I}_{t,h}} &= \sum_{s\in\mathcal{T}_t} \delta_{t,h,s}\,\mathbf{k}^{I}_{s}, \\
\frac{\partial \mathcal{L}^{I}}{\partial \mathbf{k}^{I}_{s}} &= \sum_{t\,:\,s\in\mathcal{T}_t}\ \sum_{h=1}^{H_I} \delta_{t,h,s}\,\mathbf{q}^{I}_{t,h}.
\end{align}
Every sum runs over the $k$ selected entries of a query. The backward pass thus needs $z_{t,h,s}$ only for $s\in\mathcal{T}_t$, or $kH_I$ per-head scores per query. The forward scan instead computes all $LH_I$. \Sparsereplay recomputes exactly these scores (\S\ref{sec:replay}). The indexer's inputs are detached. The gradients thus reach only the indexer's projection weights.

\paragraph{Sparse attention.}
In MLA, head $h$ of query $t$ attends to the selected latents with logits $\ell_{t,h,s} = \mathbf{q}_{t,h}^{\top}\mathbf{c}_s/\sqrt{d}$ for $s\in\mathcal{T}_t$. Together with its positional part, $\mathbf{c}_s$ serves as the key. Its latent part $\mathbf{c}^{V}_s$ serves as the value. Let $a_{t,h,s} = \mathrm{softmax}_{s\in\mathcal{T}_t}(\ell_{t,h,s})$ be the attention weights and $\mathbf{o}_{t,h} = \sum_{s\in\mathcal{T}_t} a_{t,h,s}\,\mathbf{c}^{V}_s$ the output. The standard attention backward~\cite{dao2022flashattention} restricted to $\mathcal{T}_t$ gives
\begin{align}
\mathrm{d}\ell_{t,h,s} &= a_{t,h,s}\big(\mathrm{d}\mathbf{o}_{t,h}^{\top}\mathbf{c}^{V}_s - \mathrm{d}\mathbf{o}_{t,h}^{\top}\mathbf{o}_{t,h}\big), \\
\mathrm{d}\mathbf{q}_{t,h} &= \tfrac{1}{\sqrt{d}} \sum_{s\in\mathcal{T}_t} \mathrm{d}\ell_{t,h,s}\,\mathbf{c}_s, \\
\mathrm{d}\mathbf{c}_s &= \sum_{t\,:\,s\in\mathcal{T}_t}\ \sum_{h=1}^{H} \Big(\tfrac{1}{\sqrt{d}}\,\mathrm{d}\ell_{t,h,s}\,\mathbf{q}_{t,h} + a_{t,h,s}\,\mathrm{d}\mathbf{o}_{t,h}\Big).
\end{align}
The first term of $\mathrm{d}\mathbf{c}_s$ comes from $\mathbf{c}_s$ as a key. The second comes from $\mathbf{c}_s$ as a value and lands on $\mathbf{c}^{V}_s$. The target $\mathbf{p}_t$ averages $a_{t,h,s}$ over heads and is detached. The indexer loss thus sends no gradient into sparse attention.

\paragraph{Context parallelism.}
Under CP, $\mathbf{k}^{I}_{s}$ and $\mathbf{c}_s$ are all-gathered. Each GPU thus computes partial $\partial\mathcal{L}^{I}/\partial\mathbf{k}^{I}_{s}$ and $\mathrm{d}\mathbf{c}_s$ from its own $n$ queries for every position $s$. The backward reduce-scatter of \S\ref{sec:bg-training} sums these partial gradients. Each GPU then receives the gradients of its own positions.

\section{\flashtopk Implementation}
\label{app:flashtopk}

\begin{figure}[t]
\centering
\includegraphics[width=\columnwidth]{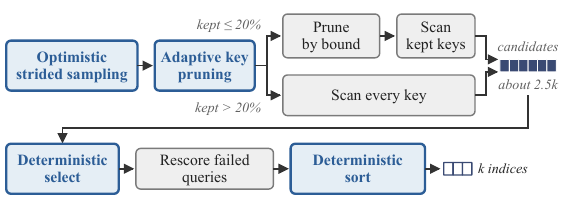}
\caption{One selection call of \flashtopk. All steps run on the GPU without host synchronization.}
\label{fig:flashtopk-flow}
\end{figure}

Figure~\ref{fig:flashtopk-flow} shows one selection call of \flashtopk. Optimistic strided sampling first sets each query's threshold (\S\ref{sec:selection}). Adaptive key pruning then estimates the share of keys that DoSA's bound keeps. If this share is at most 20\%, \flashtopk prunes by the bound and scans only the kept keys. Otherwise, it scans every key. Either scan writes each query's passing keys to its 16K-slot candidate buffer. About $2.5k$ keys pass on average. The deterministic select then keeps each query's $k$ best candidates and breaks ties at the $k$-th score by position. Recovery rescores the failed queries. The deterministic sort finally orders the $k$ indices by score, then by position. Each branch reads a flag on the GPU. No step thus synchronizes with the host.

\paragraph{Optimistic strided sampling.} The scan kernel scores the sampled keys of \S\ref{sec:selection} with its own arithmetic. A sample's score thus has the same bits as in the scan. Each query reads only the samples inside its window. \flashtopk keeps LiteTopK's bucketed threshold. Its 256 buckets span the range of a query's sample scores.

\paragraph{Adaptive key pruning.} \flashtopk estimates the kept share on one query in 16 against 256 sampled keys. Every call launches both paths. The other path sees empty windows and returns at once.

\paragraph{Prune by bound.} \flashtopk adapts DoSA's upper bound on the score of key $\mathbf{k}$ for query $t$,
\[
U = \textstyle\sum_{j \in R} \tilde{w}_j\, \mathrm{ReLU}(\mathbf{q}_j \cdot \mathbf{k}) - \mathrm{ReLU}(\mathbf{a} \cdot \mathbf{k}) + E\,\lVert \mathbf{k} \rVert.
\]
The set $R$ holds the 14 positive heads with the largest $w_h \lVert \mathbf{q}_h \rVert$. Each other positive head is projected onto its most aligned head in $R$. Its residual joins the allowance $E$. The allowance also covers rounding. The aggregate $\mathbf{a}$ sums the negative heads active on at least half of the keys. These heads can only lower the score. A key with a bound below the threshold thus cannot pass it. Each query's bound takes 16 columns, 14 for the retained heads and two for the negative aggregate. The pruning kernel takes 16 queries per tile. It writes one bit per key for each group of 8 queries.

\paragraph{Scan kept keys.} The scan locates the kept keys with a prefix sum and a binary search. It packs them into dense tiles and scores them with the same arithmetic as the scan of every key.

\paragraph{Rescore failed queries.} A check kernel marks a query as failed if its candidates overflowed the buffer or number fewer than $\min(k, L_t)$. Recovery kernels then rescore up to 64 failed queries over their windows and reselect them in place. They run after every selection with a fixed grid. Each returns at once when no query failed. With more than 64 failures, every query is rescored. This never occurred in our runs. At 1M, one failed query adds 0.3\% to the selection time, and 64 add 2.6\%.

\paragraph{Deterministic select and sort.} The scans write candidates with atomics. Their order thus varies between runs. Sometimes more candidates tie at the $k$-th score than there are places left. The select kernel then runs a radix select over the positions of just these tied keys. It keeps the tied keys with the smallest positions. The selected set thus depends only on the scores. A radix sort then orders each query's $k$ indices by score and breaks ties by position (\S\ref{sec:selection}). At 1M, the sort adds 0.8\% to the selection time.

\section{Costs of Selection Recomputation}
\label{app:replay-costs}

Table~\ref{tab:replay-costs} gives the cost of every policy in Figure~\ref{fig:replay-pareto}, in general and for GLM-5.1.

\begin{table}[ht]
\centering
\caption{Selection recomputation costs per layer and GPU. Memory excludes the layer input. It counts 4\,B per index or score and 2\,B per indexer input element. Recomputation counts 2 ops per multiply-add of scoring and 1 per score for top-$k$. GLM-5.1 has $L$~=~1M, $n$~=~8K, $k$~=~2048, $H_I$~=~32, and $d_I$~=~128.}
\label{tab:replay-costs}
\footnotesize
\setlength{\tabcolsep}{2.5pt}
\resizebox{\columnwidth}{!}{%
\begin{tabular}{@{}llr@{\hspace{10pt}}lr@{}}
\toprule
 & \multicolumn{2}{c}{Memory (B)} & \multicolumn{2}{c}{Recomputation (ops)} \\
\cmidrule(lr){2-3}\cmidrule(l){4-5}
Policy & General & GLM-5.1 & General & GLM-5.1 \\
\midrule
\multicolumn{5}{@{}l}{\emph{Operator-level policies}} \\
No recomp. & all & all & 0 & 0 \\
Full recomp. & 0 & 0 & $2nLH_Id_I{+}nL$ & 70.4\,T \\
Keep scores & $4nL$ & 32\,Gi & $nL$ & 8.6\,G \\
Keep top-$k$ & $4nk$ & 64\,Mi & $2nLH_Id_I$ & 70.4\,T \\
Keep inputs & $2(L{+}nH_I)d_I$ & 320\,Mi & $2nLH_Id_I{+}nL$ & 70.4\,T \\
\midrule
\multicolumn{5}{@{}l}{\emph{Selection replay}} \\
\textbf{\Indexreplay} & $4nk$ & 64\,Mi & $2nkH_Id_I$ & 137\,G \\
\textbf{\Scorereplay} & $8nk$ & 128\,Mi & 0 & 0 \\
\bottomrule
\end{tabular}}
\end{table}

\end{document}